\documentclass[11pt,a4paper]{article}
\usepackage[usenames,dvipsnames]{xcolor}
\usepackage{times,latexsym}
\usepackage{url}
\usepackage[T1]{fontenc}

\usepackage[acceptedWithA]{tacl2018v2}

\usepackage{xspace,mfirstuc,tabulary}

\newif\iftaclinstructions
\taclinstructionsfalse % AUTHORS: do NOT set this to true
\iftaclinstructions
\renewcommand{\confidential}{}
\renewcommand{\anonsubtext}{(No author info supplied here, for consistency with
TACL-submission anonymization requirements)}
\newcommand{\instr}
\fi

\iftaclpubformat % this "if" is set by the choice of options

\else

\fi

\usepackage{hyperref}
\usepackage{url}
\usepackage{booktabs}       % professional-quality tables
\usepackage{nicefrac}       % compact symbols for 1/2, etc.
\usepackage{microtype}      % microtypography
\usepackage{pifont}

\usepackage{enumitem}

\usepackage{wrapfig}
\usepackage{changepage}
\usepackage{amsmath}
\usepackage{amsthm}
\usepackage{amssymb}
\usepackage{amsfonts,eucal,amsbsy}
\usepackage{mathtools}
\usepackage{multirow}
\usepackage[normalem]{ulem}
\usepackage{natbib}
\usepackage{sidecap}

\usepackage{subfigure}

\newcommand{\ignore}[1]{}
\newcommand{\vect}[1]{\mathbf{#1}}

\newcommand{\revision}[1]{{#1}}

\title{Adaptive Semiparametric Language Models}

\author{
 Dani Yogatama, Cyprien de Masson d'Autume, Lingpeng Kong \\
 DeepMind \\
 London, United Kingdom \\
 {\sf \{dyogatama,cyprien,lingpenk\}@google.com} \\
}

\date{}

\begin{document}
\maketitle
\begin{abstract}
We present a language model that combines 
a large parametric neural network (i.e., a transformer)
with a non-parametric episodic memory component in an integrated 
architecture.
Our model uses 
extended short-term context by caching
local hidden states---similar to transformer-XL---and 
global long-term memory by
retrieving a set of nearest 
neighbor tokens at each timestep.
We design a gating function to adaptively combine 
multiple information sources to make a prediction.
This mechanism allows the model 
to use either local context, short-term memory, or 
long-term memory (or any combination of them) 
on an ad hoc basis depending on the context.
Experiments on word-based and character-based 
language modeling datasets
demonstrate the efficacy of our proposed method 
compared to strong baselines.
\end{abstract}

\section{Introduction}
Human language processing is facilitated by
complex systems interacting together. A
core component that enables such a process is human memory.
Memory in humans consists of specialized systems,
which forms a basis for
intelligent behaviors \citep{tulving, rolls2000, eichenbaum2012memory}.
For language processing, working (short-term) memory is 
a temporary storage that can be
used to comprehend sentences and 
follow conversations. Episodic (long-term) memory stores individual experience 
and events. Semantic memory stores facts and knowledge about words
and concepts.\footnote{We refer readers to \citet{nematzadeh} for 
discussions on human and artificial language processing memory systems.}

In artificial language processing systems (e.g., language models), 
a popular approach to
design a better model
is by encoding all of the desired knowledge (e.g., to produce 
grammatical sentences, process long text,
remember events, etc.) in the weights of a large parametric 
neural network via end-to-end training.
We see an increasingly larger transformer become
a better language model \citep{gpt,gpt2,megatron,gpt3}. 
In this \emph{scale} approach, the knowledge is 
implicitly represented in the weights of a parametric neural network, and 
it is not straightforward to interpret whether 
a model contains a particular knowledge without
asking the model to produce a response---e.g., via a cloze-style question \citep{lmiskb} 
or a prompt \citep{gpt3}.

An alternative strategy
is to design a \emph{modular} architecture that
separates memory storage and computational processing, where each module has
a clear purpose.
Recent progress in memory-augmented neural networks has given rise to
many variants of memory-augmented transformer language models
that fall under this category.
For example, attempts to incorporate
extended local context to a neural network---such as those 
found in neural cache \citep{grave}, transformer-XL \citep{txl}
compressive transformer \citep{compresstrans}, performers \citep{performer}, \revision{longformer \citep{longformer}, and reformer \citep{reformer}}---can 
be seen as 
models of working memory.
Models of episodic memory
include $k$NN-LM \citep{knnlm}
and architectures that are designed for more complicated tasks 
such as question answering \citep{demasson, realm} and machine translation \citep{nnmt}.
In
machine learning and natural language processing,
memory-augmented neural networks is used to refer to all types of memory systems.

In this paper, \revision{inspired by the modular design of human memory systems, 
we present a language model architecture (\textsc{Spalm})
with storage modules that resemble working and episodic memory systems,
which we combine with a large parametric neural network that
is responsible for computation (\S{\ref{sec:model}})}.
Our hypothesis is that encouraging
each component to focus on a specific function (e.g.,
storing long-term information, capturing extended context, 
modeling local information)
facilitates easier training that produces an overall better language model.\footnote{\revision{We note that \textsc{Spalm} is not intended to be a model of human language processing system. We merely take inspirations from human memory systems to design a better artificial language model.}}

Specifically, we follow transformer-XL \citep{txl} to 
capture extended context
by caching hidden states in a temporary short-term memory. For long-term context,
we use a persistent key-value database and perform
sparse retrieval with (approximate) $k$-nearest neighbors.
In contrast to previous language models that either interpolate output probabilities
\citep{smerity,grave,knnlm,bertknn} or use input concatenation \citep{realm,megatronctrl}
to combine information
from different sources, we design
a context-dependent gating mechanism
to incorporate local, extended, and global context.
We discuss similarities and differences to related work in \S{\ref{sec:relwork}}.

In language modeling, many tokens can be
predicted from their local context without 
requiring long-term information.
Our model can adaptively decide whether the current (local)
context is enough, or whether it needs to use information from the short-term
and/or long-term memory.

In \S{\ref{sec:experiments}}, we compare \textsc{Spalm} with strong baselines---including transformer-XL and $k$NN-LM---on word-based and character-based
language modeling.
Our positive results establish the benefit of the proposed architecture.
They also indicate the generality of our approach
and its potential applicability to other sequence modeling tasks.

We analyze how \textsc{Spalm} uses long vs. short-term context 
(\S{\ref{sec:analysis}}) to better understand how the model operates
when making predictions.
We conclude by discussing limitations and
future directions (\S{\ref{sec:discussion}}).

\section{Model}
\label{sec:model}
\begin{figure}[t]
\centering
\includegraphics[scale=0.35]{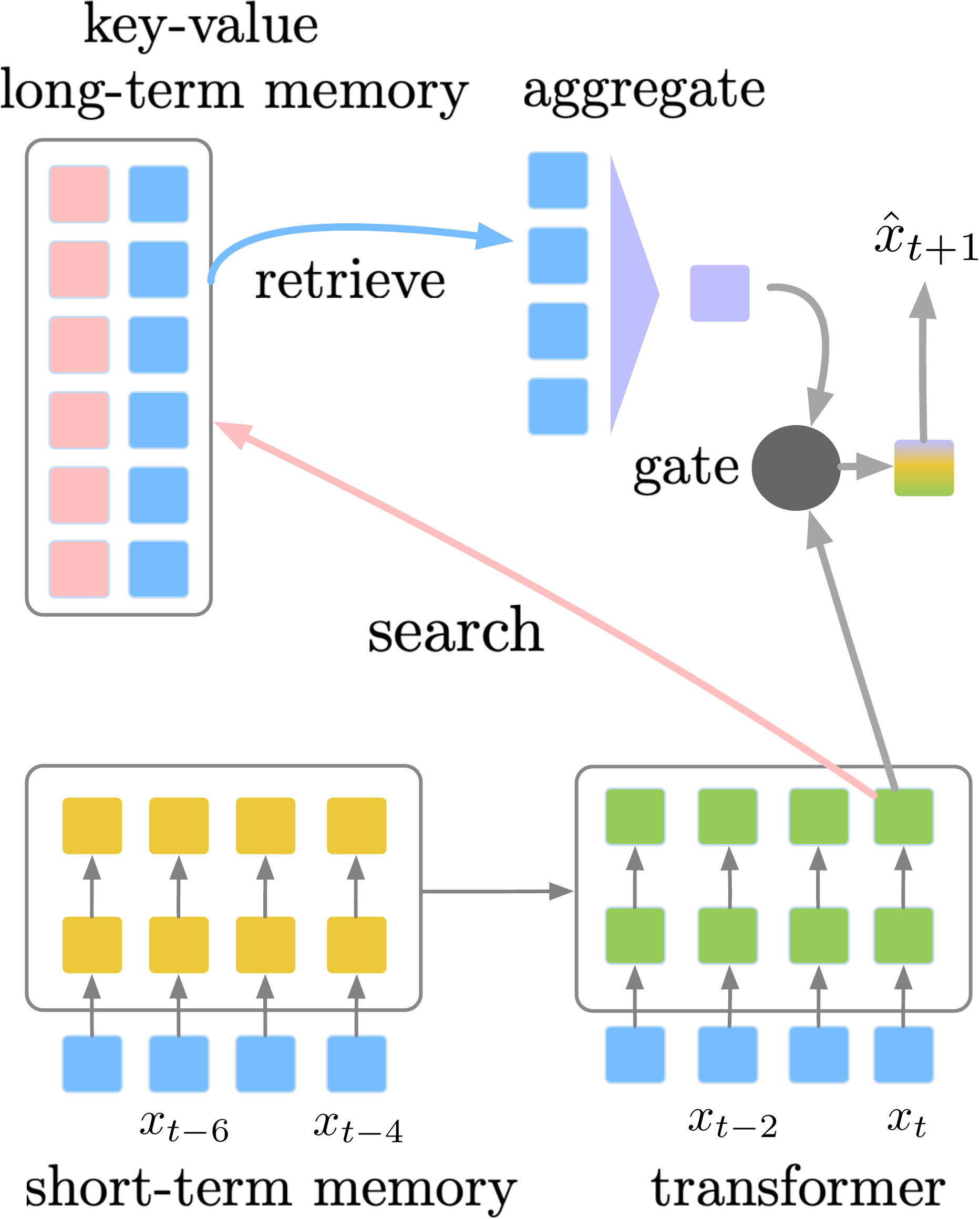}
\caption{Our language model architecture
has three main components: (i) a transformer
that processes the current local context, (ii)
a short-term memory module which stores hidden
states from an extended context, (iii)
and a key-value (hidden state-output token) 
database that stores compressed long-term context.
At each timestep, our model combines the current context and
short-term memory with a mechanism similar to transformer-XL.
It then retrieves a set of past output tokens that are
used in a similar context from the long-term memory module.
These past output tokens are then encoded and aggregated
to a single vector that represents long-term information.
We use a context-dependent gate to combine information
from multiple sources for making a final prediction.
\label{fig:model}}
\end{figure}

We consider a language model that takes as input a sequence of words
$\boldsymbol{x}_{\leq t} = \{x_0, \ldots, x_{t}\}$ and outputs a probability 
distribution of the next word $p(x_{t+1} \mid \ \boldsymbol{x}_{\leq t}; \vect{W})$.
Given a corpus of $T$ words, the log likelihood of the corpus is:
\begin{align*}
\mathcal{L} = \sum_{t=0}^{T} \log p(x_{t+1} \mid \ \boldsymbol{x}_{\leq t}; \vect{W}),
\end{align*}
where $x_0$ is the start of sentence symbol.

\textsc{Spalm} consists of three main components: (i) a large parametric neural network 
in the form of a transformer to process local context, 
(ii) a short-term memory to store extended context,
and (ii) a non-parametric episodic
memory module that stores information from long-term context.
We integrate these components in a single architecture with a
gating mechanism. Figure~\ref{fig:model} shows an illustration of our model,
which we discuss in details below.

\subsection{Base model}
We use transformer \citep{transformer} as our base model.
Given the input sequence $\boldsymbol{x}_{\leq t}$, transformer performs 
multiple layers of self-attention between every pair of 
tokens in the input sequence 
to produce token representations. 

A core limitation of transformer is that its computational
complexity is quadratic in the input sequence length.
As a result, instead of considering all previous tokens $\boldsymbol{x}_{\leq t}$,
transformer truncates the input to be the most recent $N$ words 
$\boldsymbol{\tilde{x}}_{\leq t} = \{x_{t-N+1},\ldots,{x_{t}}\}$
and only operates on this fixed-length window in practice.
A large transformer, no matter
how many parameters it has, 
is limited by the input sequence length.

\subsection{Short-term memory}

We use transformer-XL \citep{txl} as our working memory model.
Given the current context $\boldsymbol{\tilde{x}}_{<t}$, denote the extended
context of length $M$ 
by $\boldsymbol{\tilde{x}}_{\leq t-N} = \{x_{t-N-M+1},\ldots,{x_{t-N}}\}$. In other words,
the extended context is the $M$ tokens prior to the current context.
In transformer-XL, hidden states for $\boldsymbol{\tilde{x}}_{\leq t-N}$ (obtained from a
previous computation when predicting $x_{t-N+1}$) are cached. They are then used as
additional states that can be attended to during the forward pass when computing
hidden states for the current context $\boldsymbol{\tilde{x}}_{\leq t}$, 
but the values of the states are not updated during the backward pass to save
computation time. 

Formally, denote the hidden state
for $x_t$ at layer $r$ by $\vect{h}_t^r$.
Denote the hidden states associated with the current (truncated) 
context $\boldsymbol{\tilde{x}}_{<t}$ by $\vect{H}^{r} = 
[\vect{h}^{r}_{t-N}, \ldots \vect{h}^{r}_{t}]$ and
the hidden states associated with
the extended context $\boldsymbol{\tilde{x}}_{<t-N}$ by $\vect{E}^{r} = 
[\textsc{Sg}(\vect{h}^{r}_{t-N-M+1}), \ldots \textsc{Sg}(\vect{h}^{r}_{t-N})]$,
where $\textsc{Sg}$ is the stop gradient function.
Together, $\vect{H}^r$ and $\vect{E}^r$ are used as an input to an attention function 
(with relative positional encodings)
where each vector is transformed
into a key, value, query triplet 
which are used to produce $\vect{H}^{r+1}$ (i.e., hidden states for the next layer).

Note that while transformer-XL extends the context
window, the extra information is still ``local'' with respect to the sequence.

\subsection{Long-term memory}
We design a long-term episodic memory module 
that allows our language model
to retrieve ``global'' information. 
The long-term memory module is implemented
as a key-value database. The key is a vector representation of a context $\boldsymbol{\tilde{x}}_{\leq i}$ (i.e., 
we compress $\boldsymbol{\tilde{x}}_{\leq i}$ into
a vector). Each context is paired with 
the output token for that context $x_{i+1}$, which is stored as the value.
In our experiments, we store 
a key-value entry for each context-token pair in the \emph{training} corpus,
so the number of entries is equal to the number of tokens in the training corpus.

There are many choices that can be used for the key representation, which we 
denote by $\vect{d}_i$.
For example, we can 
use $\vect{h}^R_{i}$
or a separate pretrained encoder such as BERT \citep{bert}.
We pretrain a vanilla transformer language model and
use the final-layer hidden state for $\vect{d}_i$.

For predicting a new token $x_{t+1}$ given $\boldsymbol{\tilde{x}}_{\leq t}$, we first obtain $\vect{d}_{t}$
from the separate pretrained language model.
We then use $\vect{d}_{t}$ to do a $k$-nearest neighbor search on the database.
Since $\vect{d}_{t}$ is a contextual representation, this search finds
contexts that are similar to $\boldsymbol{\tilde{x}}_{<t}$ in the database.
For the top $k$ such contexts, we retrieve the values associated with those contexts, 
which are the output (next) tokens when those contexts are encountered in the past.
Denote the output tokens retrieved from the database by $y_1, \ldots y_K$.

For each $y_k$, we create a vector representation $\vect{y}_k$ by using the same word embedding
matrix that is used in our base model. We then combine the long-term memory
information obtain from the database with the extended local context with a gating
mechanism as follows:
\begin{align*}
&\vect{m}_t = \sum_{k=1}^K \frac{\exp\vect{y}_k^{\top}\vect{h}_{t}^R}{\sum_{j=1}^K \exp\vect{y}_j^{\top}\vect{h}_{t}^R} \vect{y}_k \\
&\vect{g}_t = \sigma(\vect{w}_g^{\top}\vect{h}_{t}^R)\\
&\vect{z}_t = (1-\vect{g}_t) \odot \vect{m}_{t} + \vect{g}_t \odot \vect{h}_{t}^R\\
&p(x_{t+1}\mid\boldsymbol{x}_{\leq t}) = \text{softmax}(\vect{z}_t; \vect{W}),
\end{align*}
where $\vect{w}_g$ is a parameter vector, $\sigma$ is the sigmoid function, and $\vect{W}$ is the word embedding matrix that
is shared for input and output word embeddings \citep{inan}.\footnote{In a preliminary experiment, we incorporate the nearest neighbor distance
as a bias term in the computation of $\vect{m}_t$. However, this does not improve performance, so we use the above equation in the final model.}

In the above formulation, we first aggregate information from $y_1, \ldots y_K$ 
with a simple attention mechanism using $\vect{h}_{t}^R$ as the attention query.\footnote{\revision{It is possible to first transform 
$\vect{h}_{t}^R$ (e.g., by doing a linear projection) before using it as an attention query.
We choose an untransformed version in our experiments to minimize the number of new parameters in \textsc{Spalm}. We leave explorations on the best transformation of $\vect{h}_{t}^R$ to future work.}}
We then use a context-dependent gate 
$\vect{g}_t$ that decides how much the model needs to use local information 
($\vect{h}_{t}^R$) versus long-term information ($\vect{m}_{t}$) for making the current
prediction based on the current context. Note that given the database,
the only additional parameter that needs to be trained is $\vect{w}_g$.
The result is a language model that is able to rely on short-term context 
for ``easy'' predictions while using long-term context for ``hard'' predictions
by adaptively combining short-term and long-term memory at the architectural level.

\subsection{Training details}
As discussed previously, we first train a standard transformer language model and 
use it as an encoder to compute key representations $\vect{d}_i$ 
for the episodic memory database.
Since our training datasets contain hundreds of millions of tokens, for computational
considerations, we do not update the key representations when training the
overall model.
This allows us to fix the set of nearest neighbors for each token, making training
of the overall model to be almost as fast as a vanilla transformer-XL in
terms of wall-clock time after we precompute neighbors for each token.
The value encoder, on the other hand, is updated during
training since we use the word embedding matrix to represent $\vect{y}_k$.

$k$-nearest neighbors on hundreds of millions of tokens can be computationally expensive.
We use the publicly available ScANN\footnote{\url{https://github.com/google-research/google-research/tree/master/scann}} \citep{scann} to do this efficiently, which is a 
quantization-based technique to do fast and accurate maximum inner product search.

We note that it is conceptually possible to train 
all components of our model in an end-to-end manner.
However, we leave end-to-end training to future work.
In addition, while it is possible to continually grow the
long-term memory module by storing new tokens from evaluation data, we choose 
to do a static evaluation. Therefore, we do not compare with 
dynamic evaluation models \citep{dynamicevaluation,dyneval2,unboundedcache} which adapt language models to evaluation data.

We next discuss comparisons to existing nearest neighbor and cache language models.

\section{Comparisons to previous work}
\label{sec:relwork}
\paragraph{$k$NN-LM.} There are several language models that are related to our proposed method. The closest one is $k$NN-LM \citep{knnlm}, which is another language model
that is augmented with a nearest neighbor retrieval mechanism.
$k$NN-LM is an ensemble technique that is designed to be used only at evaluation time.
In $k$NN-LM, a pretrained language model (e.g., a transformer) is combined
with another retrieval-based language model by interpolating their probabilities:
$p(x_{t+1}\mid\boldsymbol{x}_{\leq t}) = \lambda p_{\text{LM}}(x_{t+1}\mid\boldsymbol{x}_{\leq t}) + (1-\lambda) p_{\text{$k$NN}}(x_{t+1}\mid\boldsymbol{x}_{\leq t})$.
The interpolation weight $\lambda$ is tuned at the corpus level on a development set.

While this post hoc integration method used by $k$NN-LM has its merit (e.g., very practical,
fast to incorporate to any model since it does not require additional training), 
our focus is on designing a model that combines short-term and long-term memory at the architecture level. Our motivation is twofold. First, interpolating
the language model weights at the corpus level forces the model
to use the same interpolation weight $\lambda$ for $p_{\text{LM}}$ and $p_{\text{$k$NN}}$
for each token in the corpus. It cannot adaptively combine 
short-term and long-term information at the token level based on the context.
In addition, $\lambda$ needs to be tuned on an extra development set.\footnote{\revision{We note that it is possible to incorporate 
this interpolation technique during the training phase of a language model as well
to avoid having to tune $\lambda$ on a development set.
For example, \citet{neubigdyer} shows how to train a mixture of experts language models,
where the mixture weights are inferred. However, the efficacy of this approach as
a memory-augmented language model has not been explored.}}
\textsc{Spalm}, on the other hand, is able
to adjust the weights placed on $\vect{m}_t$ and $\vect{h}_{t}^R$ when constructing
$\vect{z}_t$ differently for different tokens.
Second, we believe that integration of different memory 
modules at the architectural level
is a more natural approach that could help pave the way for applications
with other memory sources (e.g., knowledge bases, images, videos)---where
the memory output is not in the same space
as the prediction output (i.e., words) \revision{and an interpolation technique cannot be used}.

We compare with $k$NN-LM in our experiments. Since interpolating model
probabilities is 
an ensembling technique that is independent of the architecture, 
we also show that our language model
can be furher ensembled with $p_{\text{$k$NN}}$ if necessary.

\paragraph{Cache-based language models and pointer networks.} Cache-based language models \citep{grave, smerity} 
store pairs of hidden states and output tokens from previously seen tokens (within
a limited context length)
in a cache. The best variant of the method uses an interpolation (ensemble) method similar to
$k$NN-LM to combine information from the cache and the backbone language model.
This class of models temporarily stores $M$ 
past hidden states (typically, in the order of thousands), 
so it is a working-memory model as opposed to long-term memory.
\revision{In addition, they also rely on interpolating probabilities of a backbone language
model and a cache component (similar to $k$NN-LN when the cache size is unbounded).}

\paragraph{Other retrieval augmented methods.}
\revision{An early version of a neural language model that includes a retrieval component is presented in \citet{guu2018}. They follow a retrieve-then-edit approach to generate a
sentence, which requires approximating an expectation over an edit prior.}

Outside language modeling, there are several recent retrieval-augmented methods
that have been used for question answering \citep{demasson,realm,xiangpaper,bertknn},
controllable generation \citep{megatronctrl}, machine translation \citep{bapna,nnmt}, \revision{and
one-shot learning \citep{rareevent}}. 
These methods share some similarities with our proposed model since
it involves a retrieval component. However, the difference in the 
downstream tasks (language modeling vs. question answering vs. machine
translation), results in different items that are stored in and retrieved from 
the key-value database.
For example, \citet{demasson} store and retrieve question-answer pairs,
\citet{realm} have a database of passages of an article, and \citet{nnmt} use
source and target sentences.
\revision{Our gating mechanism resembles the gate 
that is used to incorporate information
from a non-parametric memory component to a machine translation model in \citet{bapna},
although the memory entries, the decoder architecture, and the downstream 
task are different.}

In addition, these models are only models of long-term memory.
Their evaluation tasks often do not need working memory because the entire
input sequence is short enough that it can be fed as an input to a transformer
as a whole.

\begin{table}[t]
\small
    \centering
    \begin{tabular}{l|r|r|r|r}
    \toprule
    \textbf{Dataset} & \textbf{\# Train} & \textbf{\# Dev} & \textbf{\# Test} & \textbf{\# Vocab}\\
    \midrule
    WikiText & 110M & 0.2M & 0.3M & 33,060 \\
    WMT & 852M & 1M & 1M & 50,259 \\
    \midrule
    enwik8 & 94M & 5.2M & 5.2M & 256 \\
    \bottomrule
    \end{tabular}
    \caption{Descriptive statistics of datasets used in our experiments. \revision{For each split, we show the number of (sub)words for WikiText and WMT and the number of characters for enwik8.}
    \label{tbl:dataset}}
\end{table}

\section{Experiments}
\label{sec:experiments}
We use word-based and character-based English language model datasets--WikiText 103, WMT, and enwik8--to evaluate our proposed method. We provide descriptive statistics in Table~\ref{tbl:dataset} and discuss each dataset in the respective section below.

\subsection{Implementation details}
We use Adam \citep{adam} as our optimizer. For word-based language modeling,
we use adaptive softmax \citep{adaptivesoftmax}. We apply dropout with a rate of 0.25. 
All models are trained on 128 Tensor Processing Units until convergence with batch size 256.

\subsection{WikiText-103}
\label{sec:wt103}
Our first dataset is WikiText-103 \citep{smerity}. 
We compare four models: vanilla transformer,
transformer-XL, $k$NN-LM, and \textsc{Spalm}.
For WikiText-103, all of our models have 
18 layers and 512 hidden dimension size with a total of 142M parameters.
We set the sequence length to 512.
For transformer-XL, we set the short-term memory length to $512$ during
training and $512$ or $3072$ at test time.
We use 4 nearest neighbors for $k$NN-LM and \textsc{Spalm} and analyze the 
effect of
varying the number
of neighbors in \S{\ref{sec:analysisnumneighobrs}}.
For $k$NN-LM, we use the transformer-XL model
to obtain $p_{\text{LM}}$, compute $p_{k\text{NN}}$
based on the nearest neighbor distance similar to \citet{knnlm}, 
and tune $\lambda$ from $\{0.05, 0.1,0.2,0.3,0.4\}$ on the development set

Table~\ref{tbl:wt103} shows perplexity on WikiText103.
Our implementation produces results that are 
in the same range as state-of-the-art numbers, demonstrating the strength of our baselines.
Transformer-XL outperforms transformer, and interpolating
the probability of transformer-XL with $k$NN (i.e., $k$NN-LM) improves the result further.
This is true both with transformer-XL (short-term) memory length of $512$
and $3072$. Comparing $k$NN-LM with $\textsc{Spalm}$, $k$NN-LM
is marginally better on the test set even though $\textsc{Spalm}$ is marginally 
better on the development set.

We observe further improvements in \textsc{Spalm} by
interpolating its output probability with the output probability from $p_{k\text{NN}}$
which is used by $k$NN-LM, resulting in the best model with
a perplexity of 17.6. 
We find this interesting since \textsc{Spalm} and $p_{k\text{NN}}$
uses the exact same four neighbors for each token.
It indicates that there are some 
complementary benefits
in incorporating long-term memory into training and interpolating probabilities at test time.

\begin{table}[ht]
\small
    \centering
    \begin{tabular}{l|l|r|r|r}
    \toprule
    &\textbf{Model} & \textbf{\# Params} & \textbf{Dev} & \textbf{Test}\\
    \midrule
    &Transformer-XL$^\text{a}$ & 257M & - & 18.3 \\
    &Adaptive Input$^\text{b}$ & 247M & 18.0 & 18.7 \\
    &Compressive$^\text{c}$ & 257M & 16.0 & 17.1\\
    &$k$NN-LM$^\text{d}$ & 247M & 16.1 & 16.1 \\
    \midrule
    \multirow{5}{*}{\rotatebox[origin=c]{90}{$M=512$}}&Transformer & 142M & 20.8 & 21.8 \\
    &Transformer-XL & 142M & 18.7 & 19.6 \\
    &$k$NN-LM & 142M & 18.1 & 18.5\\
    &\textsc{Spalm} & 142M & 17.9 & 18.8 \\
    &\quad$\hookrightarrow$ + $k$NN & & 17.6 & 18.0\\
    \midrule
    \multirow{4}{*}{\rotatebox[origin=c]{90}{$M=3072$}}&Transformer-XL & 142M & 18.3 & 19.1\\
    &$k$NN-LM & 142M & 17.7 & 18.0\\
    &\textsc{Spalm} & 142M & 17.4 & 18.3\\
    &\quad$\hookrightarrow$ + $k$NN & & 17.2 & \textbf{17.6}\\
    \bottomrule
    \end{tabular}
    \caption{Perplexity on WikiText-103. The top rows contain results
    taken from other papers: (a) transformer-XL \citep{txl}, (b) adaptive input
    embeddings \citep{adaptiveinput}, (c) compressive transformer \citep{compresstrans},
    and (d) $k$NN-LM \citep{knnlm}. \revision{The (log likelihood) difference between the best model (\textsc{Spalm + $k$NN}) and transformer-XL on the test set is statistically significant (Wilcoxon signed-rank test, $p < 0.05$).}
    \label{tbl:wt103}}
\end{table}

\subsection{WMT}
\label{sec:wmt}
In the second experiment, our goal is to evaluate on a much larger
dataset. We construct a language modeling dataset from the English portion of 
the WMT 2019 dataset, publicly available at \url{http://www.statmt.org/wmt19/}.
WMT contains news articles from different months. We use
articles from January to October for training, a portion of articles in November 
for development, and a portion of articles in December for test.\footnote{We sample 
articles written in November and December in chronological order 
to create development
and test sets of approximately 1 million tokens (there are almost 100 million tokens if we use all of the articles in each month).} The resulting WMT dataset is approximately ten times larger than 
the WikiText-103 dataset.

Similar to the previous experiment, we evaluate models with 18 layers and 512 hidden dimension size with a total of 148 million parameters. 
We set the sequence length to 512, the transformer-XL short-term memory length to 512 for training and evaluation, and the number of neighbors for $\textsc{Spalm}$ and $k$NN-LM to 4.

Table~\ref{tbl:wmt} shows
results on this dataset. Consistent with the previous experiment,
$k$NN-LM outperforms transformer-XL and transformer. \textsc{Spalm}
outperforms all of them by a considerable margin on the test set. Unlike WikiText-103, 
we observe no further improvement
interpolating the probabilities of \textsc{Spalm} with $p_{k\text{NN}}$.
The results also indicate that when the distributions of
the dev and test sets can be different (e.g., articles
from different months), $k$NN-LM that relies on tuning $\lambda$ on
the dev set is more sensitive to performance discrepancy between the dev
and test sets.

\begin{table}[ht]
\small
    \centering
    \begin{tabular}{l|r|r|r}
    \toprule
    \textbf{Model} & \textbf{\# Params} & \textbf{Dev} & \textbf{Test}\\
    \midrule
    Transformer & 148M & 16.0 & 16.3\\
    Transformer-XL & 148M & 15.6 & 15.5 \\
    $k$NN-LM & 148M & 13.1 & 15.2 \\
    \textsc{Spalm} & 148M & 13.0& \textbf{14.0}\\
    \bottomrule
    \end{tabular}
    \caption{Perplexity on the WMT dataset. \revision{The (log likelihood) difference between \textsc{Spalm} and transformer-XL on the test set is statistically significant (Wilcoxon signed-rank test, $p < 0.05$).}
    \label{tbl:wmt}}
\end{table}

\begin{figure*}[ht]
\footnotesize
\begin{tabular}{lllllllllllll}
For & \textcolor{ForestGreen}{Warren} & \& & Wednesday &briefly &a &5 &billion &to & equity \\
 &\textcolor{ForestGreen}{Warren} &may &Tuesday &praised &wiping &16 &\textcolor{ForestGreen}{trillion} &\textcolor{ForestGreen}{in} & funding\\
Perhaps &\textcolor{ForestGreen}{Warren}& has &Sunday &stood &breaking &10 &billion &for & \textcolor{ForestGreen}{federal}\\
Like &\textcolor{ForestGreen}{Warren} &,& Monday &defended &using &166 &\textcolor{ForestGreen}{trillion} &\textcolor{ForestGreen}{in} & spending \\
\textcolor{RoyalBlue}{Elizabeth} &\textcolor{RoyalBlue}{Warren} &\textcolor{RoyalBlue}{on} &\textcolor{RoyalBlue}{Friday} &\textcolor{RoyalBlue}{proposed} &\textcolor{RoyalBlue}{\$} &\textcolor{RoyalBlue}{20} &\textcolor{RoyalBlue}{trillion} &\textcolor{RoyalBlue}{in} & \textcolor{RoyalBlue}{federal}\\
\\
grants & in& 10 &course &eight &. &fight &even &\textcolor{ForestGreen}{care} &for\\
funding & \textcolor{ForestGreen}{over} &\textcolor{ForestGreen}{the} &\textcolor{ForestGreen}{next}& three &. &upgrade &them & coverage & for \\
funds &\textcolor{ForestGreen}{over} &10 &\textcolor{ForestGreen}{next} &five &in &improve &American & - & \textcolor{ForestGreen}{to} \\
, &\textcolor{ForestGreen}{over} &a& \textcolor{ForestGreen}{next} &10 &, &invest &a & insurance &services\\
\textcolor{RoyalBlue}{spending}& \textcolor{RoyalBlue}{over}& \textcolor{RoyalBlue}{the}& \textcolor{RoyalBlue}{next}& \textcolor{RoyalBlue}{decade}& \textcolor{RoyalBlue}{to}& \textcolor{RoyalBlue}{provide}& \textcolor{RoyalBlue}{health} & \textcolor{RoyalBlue}{care} &\textcolor{RoyalBlue}{to}\\
\\
more &community& as& the& rates& . &\textcolor{ForestGreen}{the} & \textcolor{ForestGreen}{middle} &\textcolor{ForestGreen}{class} \\
everyone &child &, &a &\textcolor{ForestGreen}{taxes} &\textcolor{ForestGreen}{on} &\textcolor{ForestGreen}{the} & wealthy& \textcolor{ForestGreen}{class} \\
some &baby &, &co &\textcolor{ForestGreen}{taxes} & . & \textcolor{ForestGreen}{the} & \textcolor{ForestGreen}{middle} &\textcolor{ForestGreen}{class} \\
\textcolor{ForestGreen}{every} &\textcolor{ForestGreen}{American} &by &triggering &\textcolor{ForestGreen}{taxes} &\textcolor{ForestGreen}{on} &all & \textcolor{ForestGreen}{middle} &\textcolor{ForestGreen}{class} \\ 
\textcolor{RoyalBlue}{every} &\textcolor{RoyalBlue}{American} &\textcolor{RoyalBlue}{without} &\textcolor{RoyalBlue}{raising} &\textcolor{RoyalBlue}{taxes} &\textcolor{RoyalBlue}{on} &\textcolor{RoyalBlue}{the} & \textcolor{RoyalBlue}{middle} &\textcolor{RoyalBlue}{class} \\
\end{tabular}
\caption{
\revision{
A sequence of words from WMT and its four nearest neighbors at each position. We break down the sequence into four blocks. The bottom row of each block in \textcolor{RoyalBlue}{blue} represents the
original sequence, which is \texttt{Elizabeth Warren on Friday~...~the middle class}. Each row above it represents a nearest neighbor token (starting from the first neighbor at the second-bottom to the fourth neighbor at the top) that is used when predicting that particular word. We highlight matching neighbor--target words
in \textcolor{ForestGreen}{green}. We provide a more detailed discussion in \S{\ref{sec:neighborexamples}}. \label{fig:exampleswmt}}}
\end{figure*}

\subsection{enwik8}
\label{sec:ew8}
In the third experiment, we evaluate our models on
character-level language modeling.
Compared to word-level language modeling, character-level has a much smaller output space (in the order of hundreds instead of tens of thousands) and has a different characteristic in how much local vs. global contexts are needed to make a good prediction.

The enwik8 dataset \citep{enwik8} is a benchmark for character-level language modeling.
We use a 24 layer model with 512 hidden size. In total, our model 
has 100 million parameters. 
We set the sequence length to 768, the transformer-XL short-term memory length 
to 1536 for training and 4096 for evaluation.
Since character-level language models has a much smaller output space, we only retrieve
two neighbors per character.

We show the results in Table~\ref{tbl:enwik8}.
Unlike the previous two word-level language modeling 
results, $k$NN-LM underperforms transformer-XL. 
However, \textsc{Spalm} outperforms all other models.
We note that a decrease 
of 0.01 is considerable on this dataset under the BPC metric.
Similar to WMT, interpolating the probabilities
of \textsc{Spalm} with $p_{k\text{NN}}$ does
not improve performance.
These results highlight a major strength of our proposed model: uniformly
setting interpolation weights at the corpus level decreases performance (i.e., $k$NN-LM), but
allowing the model to flexibly decide when to use long-term vs. 
short-term memory is beneficial.

Since character-level and word-based language modeling are characteristically different, the success of our model on this dataset indicates its applicability to other sequence modeling problems. We leave such explorations to future work.

\begin{table}[htb]
\small
    \centering
    \begin{tabular}{l|r|r|r}
    \toprule
    \textbf{Model} & \textbf{\# Params} & \textbf{Dev} & \textbf{Test}\\
    \midrule
    18L Transformer-XL$^\text{a}$ & 88M & - & 1.03 \\
    24L Transformer-XL$^\text{a}$ & 277M & - & 0.99 \\
    Longformer$^\text{c}$ & 102M & - & 0.99 \\
    Compressive$^\text{d}$ & 277M & - & 0.97 \\
    \midrule
    Transformer & 104M & 1.07 & 1.05 \\
    Transformer-XL & 104M & 1.03 & 1.01 \\
    $k$NN-LM & 104M & 1.04 & 1.02\\
    \textsc{Spalm} & 104M & 1.02 & \textbf{1.00} \\
    \bottomrule
    \end{tabular}
    \caption{Bits per character (BPC) on enwik8. The top rows contain results
    taken from other papers: (a) transformer-XL \citep{txl}, (b) longformer \citep{longformer}, and (c) compressive transformer \citep{compresstrans}. \revision{The (log likelihood) difference between \textsc{Spalm} and transformer-XL on the test set is statistically significant (Wilcoxon signed-rank test, $p < 0.05$).}
    \label{tbl:enwik8}}
\end{table}

\begin{figure*}[ht]
\small
\begin{tabular}{lllllllllllllllllllllllllll}
U &o &\textcolor{ForestGreen}{e} &h &  &a &t &\textcolor{ForestGreen}{f} &\textcolor{ForestGreen}{o} &\textcolor{ForestGreen}{r} &\textcolor{ForestGreen}{e} & &\textcolor{ForestGreen}{t} &\textcolor{ForestGreen}{h} &i &  &d &e &n &\textcolor{ForestGreen}{i} &\textcolor{ForestGreen}{e} &\textcolor{ForestGreen}{t} &- &U &\textcolor{ForestGreen}{n} &t &\textcolor{ForestGreen}{a} \\
&h &\textcolor{ForestGreen}{e} &r &  &h &\textcolor{ForestGreen}{e} &\textcolor{ForestGreen}{f} &\textcolor{ForestGreen}{o} &\textcolor{ForestGreen}{r} &\textcolor{ForestGreen}{e} &  &\textcolor{ForestGreen}{t} &\textcolor{ForestGreen}{h} &\textcolor{ForestGreen}{e} &  &f &a &v &\textcolor{ForestGreen}{i} &\textcolor{ForestGreen}{e} &\textcolor{ForestGreen}{t} &' &U &\textcolor{ForestGreen}{n} &\textcolor{ForestGreen}{v} &\textcolor{ForestGreen}{a} \\
\textcolor{RoyalBlue}{E} &\textcolor{RoyalBlue}{v} &\textcolor{RoyalBlue}{e} &\textcolor{RoyalBlue}{n} & & \textcolor{RoyalBlue}{b}& \textcolor{RoyalBlue}{e}& \textcolor{RoyalBlue}{f}& \textcolor{RoyalBlue}{o} &\textcolor{RoyalBlue}{r} &\textcolor{RoyalBlue}{e} & &\textcolor{RoyalBlue}{t} &\textcolor{RoyalBlue}{h} &\textcolor{RoyalBlue}{e} & & \textcolor{RoyalBlue}{S}& \textcolor{RoyalBlue}{o} &\textcolor{RoyalBlue}{v} &\textcolor{RoyalBlue}{i} &\textcolor{RoyalBlue}{e} &\textcolor{RoyalBlue}{t} & &\textcolor{RoyalBlue}{i} &\textcolor{RoyalBlue}{n} &\textcolor{RoyalBlue}{v} &\textcolor{RoyalBlue}{a} \\
\\
\textcolor{ForestGreen}{s} &\textcolor{ForestGreen}{i} &\textcolor{ForestGreen}{o} &\textcolor{ForestGreen}{n} &, &b &n & &\textcolor{ForestGreen}{t} &\textcolor{ForestGreen}{h} &\textcolor{ForestGreen}{e} & &[ &\textcolor{ForestGreen}{n} &\textcolor{ForestGreen}{d} & &\textcolor{ForestGreen}{o} &\textcolor{ForestGreen}{f} &  &t &\textcolor{ForestGreen}{[} &\textcolor{ForestGreen}{1} &6 &\textcolor{ForestGreen}{7} &5 &\textcolor{ForestGreen}{]} &\textcolor{ForestGreen}{]} \\
\textcolor{ForestGreen}{s} &\textcolor{ForestGreen}{i} &\textcolor{ForestGreen}{o} &\textcolor{ForestGreen}{n} &  &\textcolor{ForestGreen}{a} &n &t &A &\textcolor{ForestGreen}{h} &\textcolor{ForestGreen}{e} & & h & \textcolor{ForestGreen}{n} &\textcolor{ForestGreen}{d} & &\textcolor{ForestGreen}{o} &\textcolor{ForestGreen}{f} &  &t &\textcolor{ForestGreen}{[} &4 &3 &9 &\textcolor{ForestGreen}{9} &\textcolor{ForestGreen}{]} &\textcolor{ForestGreen}{]} \\
\textcolor{RoyalBlue}{s} &\textcolor{RoyalBlue}{i} &\textcolor{RoyalBlue}{o} &\textcolor{RoyalBlue}{n} &  &\textcolor{RoyalBlue}{a} &\textcolor{RoyalBlue}{t} &  &\textcolor{RoyalBlue}{t} &\textcolor{RoyalBlue}{h} &\textcolor{RoyalBlue}{e} & & \textcolor{RoyalBlue}{e} &\textcolor{RoyalBlue}{n} &\textcolor{RoyalBlue}{d} & &\textcolor{RoyalBlue}{o} &\textcolor{RoyalBlue}{f} & &\textcolor{RoyalBlue}{[} &\textcolor{RoyalBlue}{[} &\textcolor{RoyalBlue}{1} &\textcolor{RoyalBlue}{9} &\textcolor{RoyalBlue}{7} &\textcolor{RoyalBlue}{9} &\textcolor{RoyalBlue}{]} &\textcolor{RoyalBlue}{]} \\
\end{tabular}
\caption{
\revision{A sequence of characters from enwik8 and its two nearest neighbors at each position. We break down the sequence into two blocks. The bottom row of each block in \textcolor{RoyalBlue}{blue} represents the
original character sequence , which is \texttt{Even before~...~[[1979]]}. The two rows above it represent the nearest neighbors (the first nearest neighbors at the second bottom row and the second nearest neighbors at the top row) that are used when predicting that particular character. We highlight matching neighbor--target characters
in \textcolor{ForestGreen}{green}. We provide a more detailed discussion in \S{\ref{sec:neighborexamples}}.
\label{fig:neighborexampleew8}}}
\end{figure*}

\begin{figure*}[htb]
\centering
... Several companies have \textcolor{RoyalBlue}{pulled their} \textcolor{ForestGreen}{advertising} \textcolor{RoyalBlue}{from} the \textcolor{RoyalBlue}{TV} show \textcolor{RoyalBlue}{following} the \textcolor{ForestGreen}{revelations} ... \\
... Liberal \textcolor{RoyalBlue}{Democrat} leader Jo \textcolor{RoyalBlue}{Swinson has} said \textcolor{RoyalBlue}{she} would work \textcolor{RoyalBlue}{with} Donald Trump in government as ...
... Additionally , \textcolor{ForestGreen}{the airline} has purchased six \textcolor{ForestGreen}{Boeing 787} - \textcolor{RoyalBlue}{9 Dream} \textcolor{ForestGreen}{liner} aircraft that are scheduled ...
\caption{
\revision{Three example sequences from the WMT test set. We highlight words where both $p_{\text{TXL}}$ and $p_{\textsc{Spalm}}$ are larger than $p_{\text{transformer}} + 0.1$ in \textcolor{ForestGreen}{green} and $p_{\textsc{Spalm}} > p_{\text{TXL}} + 0.1$ in \textcolor{RoyalBlue}{blue}. See \S{\ref{sec:outputanalysis}} for details.
\label{fig:probexamplewmt}}}
\end{figure*}

\section{Analysis}
\label{sec:analysis}
We have demonstrated the efficacy of our proposed method on three language
modeling tasks. In this section, we analyze the model to gain more insights 
into how it works.

\subsection{Examples of neighbors}
\label{sec:neighborexamples}
We inspect the neighbor tokens that are retrieved from the long-term memory
for news articles in the WMT development dataset. 
We provide
a cherry-picked example in Figure~\ref{fig:exampleswmt}. As
the model sees more tokens in a sequence, the long-term memory model becomes more accurate.
\revision{We observe interesting cases such as when predicting a named entity (e.g., \texttt{Elizabeth Warren}), 
even if the long-term memory model fails to retrieve the correct first name, 
it usually is able
to retrieve the correct last name after seeing the first name (because the entity exists in the training corpus). We observe this phenomenon in many other examples as well. We can also see that the retrieved neighbors are generally relevant even when they do not match a target word exactly---e.g., when predicting names of days, dollar amounts, time quantifiers, and common phrases.}

We next investigate neighbors on enwik8 development set (Figure~\ref{fig:neighborexampleew8}).
\revision{We observe that information from the long-term memory helps
when completing common words (e.g., \texttt{before} and \texttt{invasion}), named entities (e.g., \texttt{Soviet}), and corpus-specific formats (e.g., double square brackets).}

We note that the above examples are only provided to give a better insight into
our model. It is entirely plausible that a baseline parametric model is already
able to predict correctly from the local context. Nonetheless, directly providing this
information as a long-term context helps
our model learn better, as evident from the superior performance of \textsc{Spalm} 
on our three evaluation datasets.

\revision{
\subsection{Output analysis}
\label{sec:outputanalysis}
We search for predictions where \textsc{Spalm} significantly outperforms transformer-XL and transformer to understand when modeling local information is sufficient (i.e., vanilla transformer), when adding extended context helps (i.e., transformer-XL), and when storing long-term information is useful (i.e., \textsc{Spalm}). We show three examples from the WMT test set in Figure~\ref{fig:probexamplewmt}.
}

\revision{
While it is difficult to find consistent patterns, 
we observe that \textsc{Spalm} is generally better than both transformer and transformer-XL for predicting (completing) common phrases and named entities (that exist in the training set), especially when they are encountered for the first time and have not appeared in the extended context (e.g., \texttt{pulled their advertising from, Liberal Democrat, Jo Swinson, Boeing 787-9 Dreamliner}).} 

\revision{
On the other hand, we also see a few cases when transformer-XL outperforms \textsc{Spalm}. These are usually associated with scenarios where the same word has appeared in the extended context. While \textsc{Spalm} uses information from the extended context as well, the probability is smoothed over by information from the long-term memory, resulting in a more peaky distribution for transformer-XL.}

\subsection{Gate vectors}
\begin{figure}[h]
\centering
\includegraphics[scale=0.24]{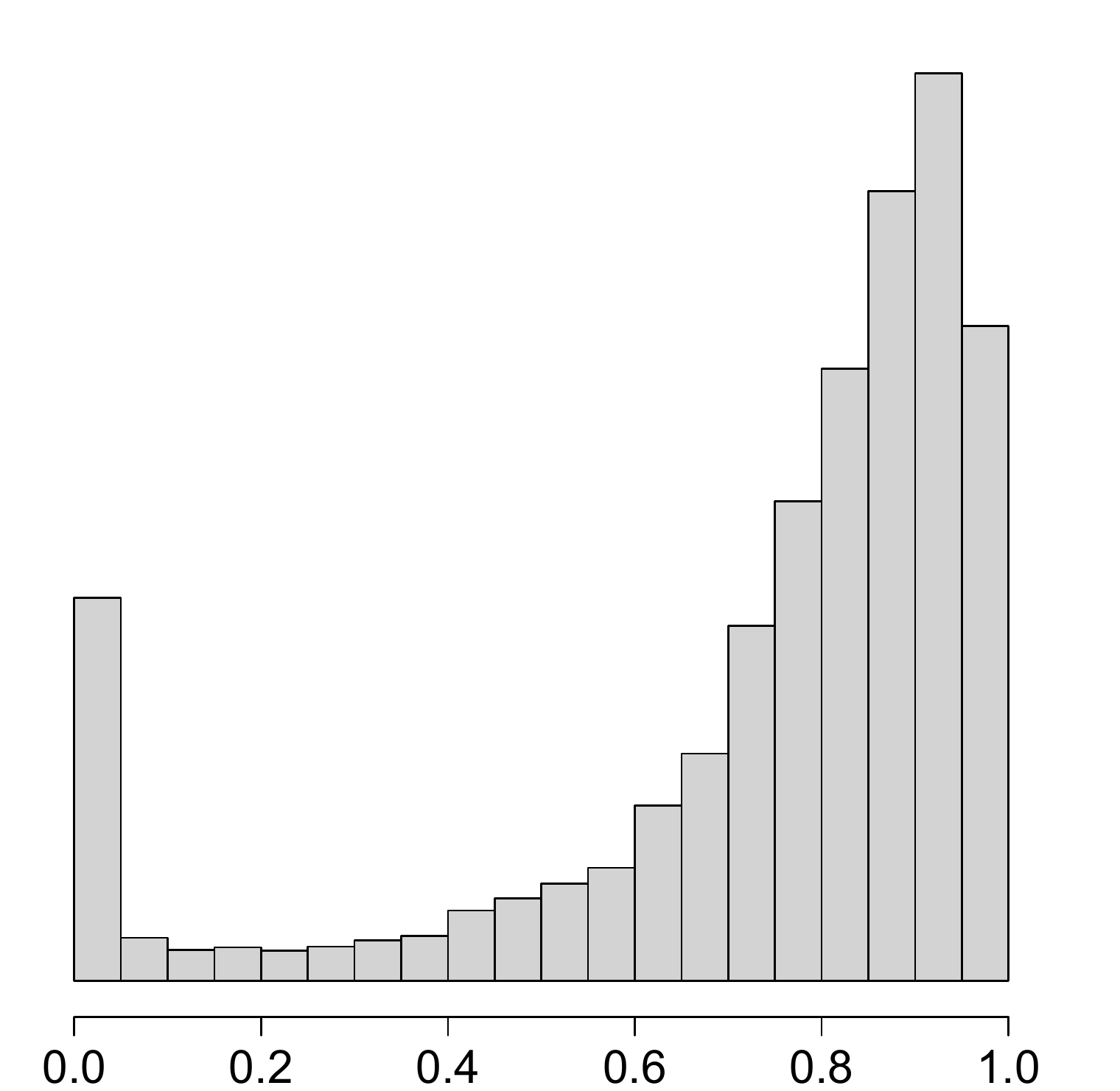}
\includegraphics[scale=0.24]{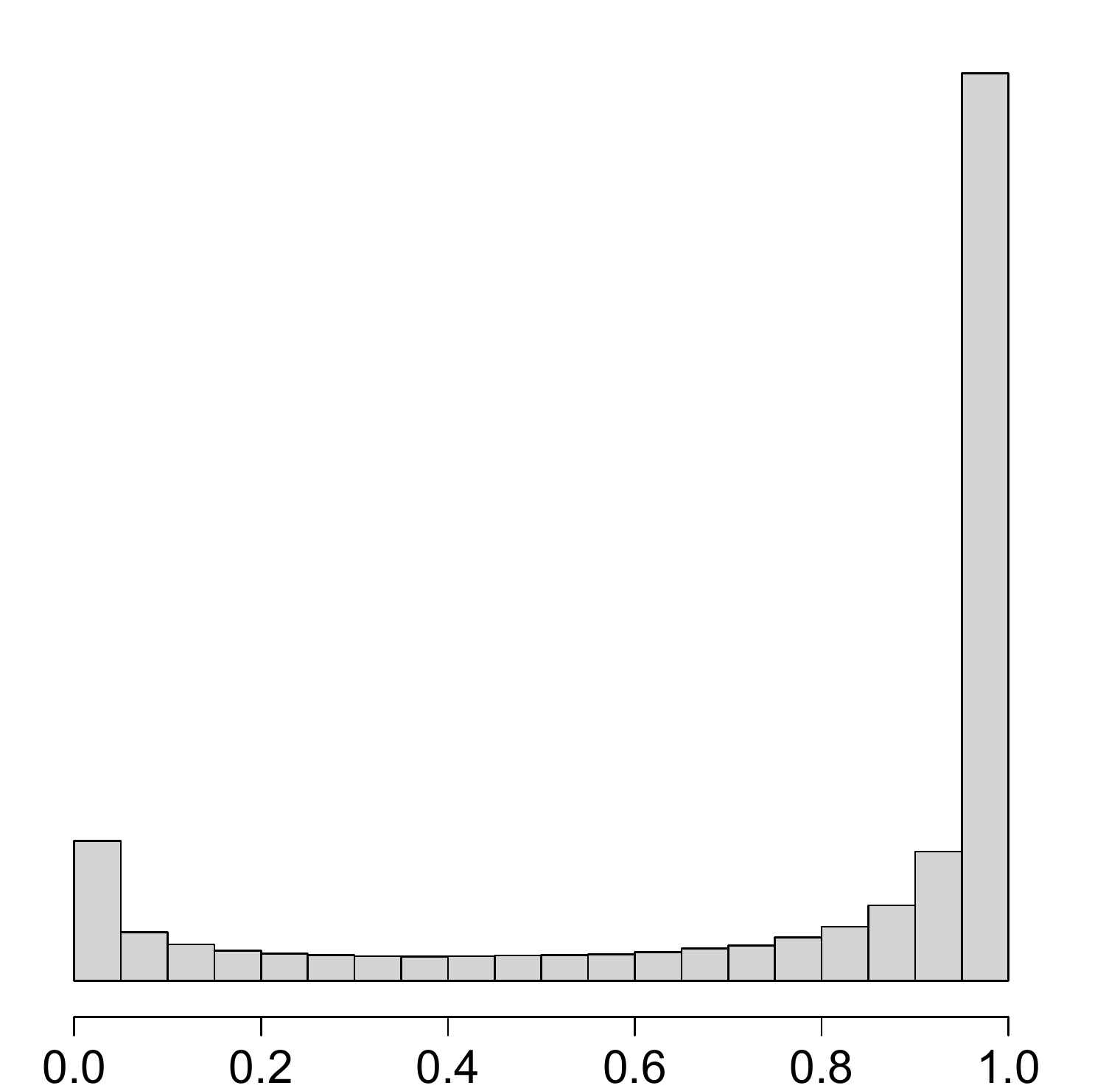}
\caption{Distributions of values of $\vect{z}$ for WMT (left) and enwik8 (right) development sets.
\label{fig:hist}}
\end{figure}

Our model has a gating mechanism to regulate information flow from
the current context, short-term, and long-term memory. We analyze 
the values of the gate for tokens in WMT and enwik8.
Figure~\ref{fig:hist} shows histograms of the distribution of gate values. 

We observe different characterstics for WMT and enwik8.
On enwik8, the gate values are concentrated around 1.
This indicates that the model relies on local context most of the time.
This can explain why $k$NN-LM does not work well on this dataset.
On WMT, the values are less concentrated around 1. This suggests
that the model uses long-term memory more than on enwik8.
\textsc{Spalm} is able to learn 
when the long-term memory is needed and when it is not in both cases.

\begin{figure*}[t]
\includegraphics[scale=0.35]{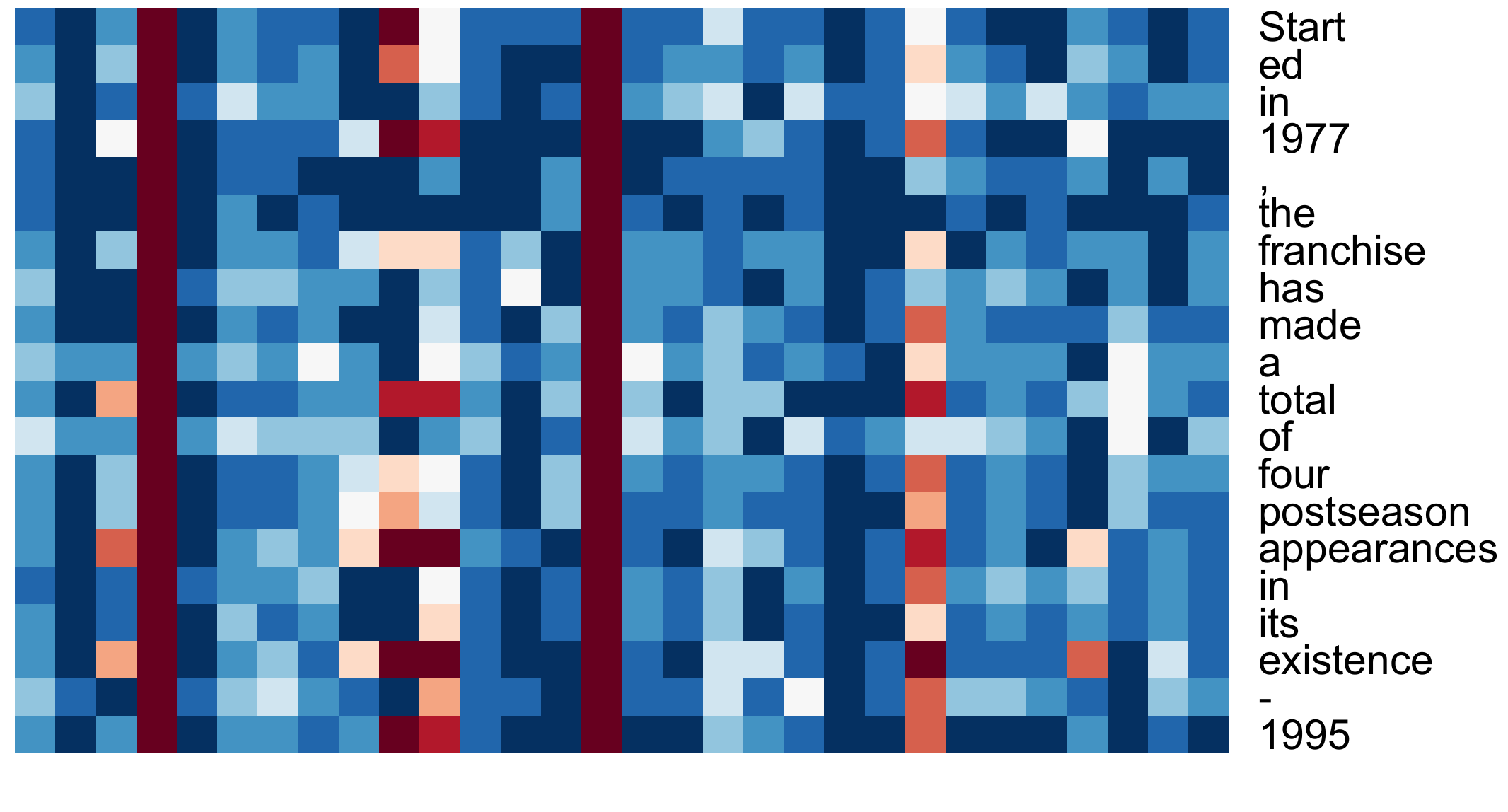}\qquad
\includegraphics[scale=0.35]{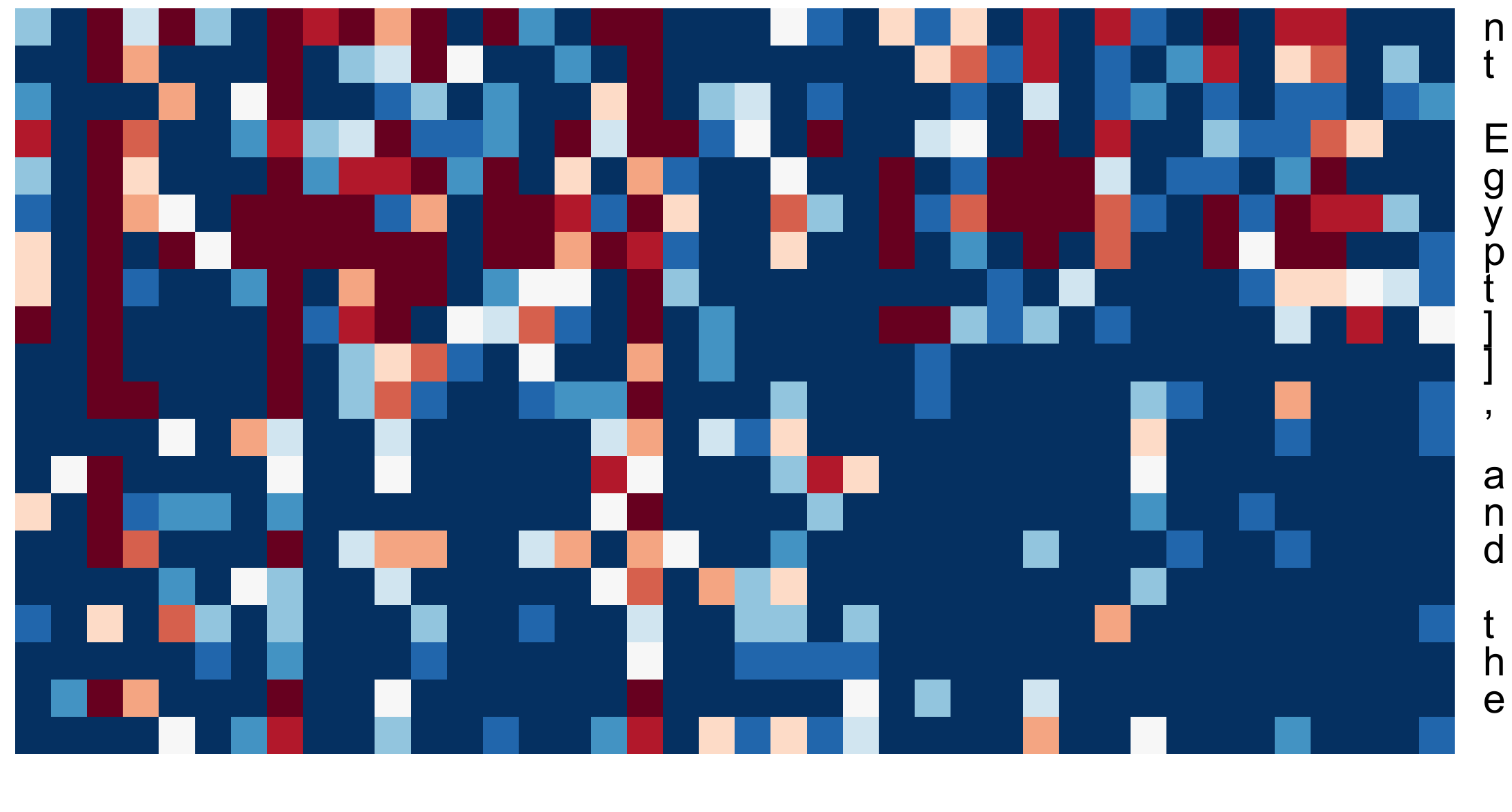}
\caption{Heatmaps of $\vect{z}$ values on a partial sequence from WMT development set (left) and enwik8 (right). Each row is a token (word or character), each colum is a dimension from $\vect{z}$.
\textcolor{blue}{blue} indicates value closer to 1.0, whereas \textcolor{red}{red} indicates value closer to 0.0. The darker the shade the closer the value is to the extreme. We see vertical patterns on WMT, indicating that these dimensions are reserved to flow information from long-term memory. Horizontal patterns on enwik8 indicates the model 
relies on long-term memory to predict a target token (e.g., when forming the word \texttt{Egypt}). The $\vect{z}$ vector has 512 dimension, we only zoom in to a small dimension subset here.
There are more horizontal and vertical patterns on both datasets as a whole.\label{fig:gates}}
\vspace{-0.5cm}
\end{figure*}

We next look into the value of the gates for a specific sequence in the development set
in Figure~\ref{fig:gates}. We note that we only show a small dimension subset
from the gate vector for readability, so we caution against drawing 
a conclusion about how the model works from this. Our goal is only to get a
better understanding of what happens when the model makes predictions.
Comparing WMT and enwik8, we see that in general on WMT the model
tends to reserve some dimensions to propagate information from the long-term memory, as indicated by vertical red lines. On enwik8, the model relies on long term information when completing a
known word such as \texttt{Egypt}, as shown by more horizontal red patterns when forming this word. For other characters, the value of the gates are closer to one, which shows that the
model relies more on local and extended short-term context.

\subsection{Number of neighbors}
\label{sec:analysisnumneighobrs}
We use four neighbors for our word-based
and two neighbors for our character-based language models.
These values are chosen from preliminary experiments on a small subset of the datasets.

We show \textsc{Spalm} perplexity on development set for WikiText-103 when we vary
the number of neighbors in Table~\ref{tbl:numneighbors}. 
We see that using one nearest neighbor is enough to
obtain good performance, with a slight advantage when we use four neighbors. 
The performance starts to degrade as we use 8 and 16 neighbors. We choose to use 
four neighbors in our
experiments since $k$NN-LM--which also uses the same set of neighbors--performs 
better with four neighbors instead of one, and we want
to keep the comparison as fair as possible.

\begin{SCtable}[\sidecaptionrelwidth][ht]
\small
\vspace{-0.2cm}
    \begin{tabular}{r|r}
    \toprule
    \textbf{\# NNs} & \textbf{Perplexity}\\
    \midrule
    1 &18.0\\
    2 &18.0\\
    4 &17.9\\
    8 &18.2\\
    16 &18.4\\
    \bottomrule
    \end{tabular}
    \caption{\textsc{Spalm} perplexity on the WikiText-103 development set with different numbers of neighbors.\label{tbl:numneighbors}}
\end{SCtable}

One notable difference between our neighbors and those that are used in $k$NN-LM \citep{knnlm}
is that we do not limit the search of the neighbors to the same token
as the current input token ($\mathbb{I}(x_i = x_t)$). While this allows the model to combine information 
from related words (not constrained to an exact match), it could introduce noise when the number of neighbors is large.

We observe that 
our representation learning model (i.e., the baseline transformer) 
is able to retrieve relevant neighbors most of the time. 
It retrieves the exact output token as the first neighbor 33\%, 44\%, and
70\% on WikiText-103, WMT and enwik8 development sets respectively.

\section{Discussion}
\label{sec:discussion}
\paragraph{Summary of contributions.}
We present a semiparametric language model (\textsc{Spalm}) that combines 
local context, short-term memory, and long-term memory to make predictions.
Experiments on word-based and character-based language models demonstrate the
benefit of our proposed method.

\paragraph{Limitations.}
The biggest limitation is the necessity to retrieve neighbors for each training token.
Such a process---even though can be fully parallelized---is time consuming.
In our epxeriments, it takes 6-8 hours to obtain neighbors for WikiText-103 and enwik8 with 1,000 CPUs and 18 hours for WMT with 9,000 CPUs.

\paragraph{Future directions.}
Our modular approach that combines multiple memory systems at the
architectural level opens up the possibility to incorporate additional
memory from other modalities (e.g., images) or structured knowledge bases.
We also envision a next-generation model that does not 
have to retrieve information
from long-term memory for every token and only does it for those
that require global context. A model that learns how to do this would save
a considerable amount of training and test time---since 
it would significantly reduce the number of
search that needs to be performed.
Our language model that integrates retrieval into training is a first step
in this direction.

\section*{Acknowledgements}
We thank the action editor (Mihai Surdeanu) and three anonymous reviewers for helpful comments on an earlier draft of this article.

\bibliographystyle{acl_natbib}
\bibliography{paper}

\end{document}